\def\mytitle   {Scalable Real-Time Vehicle Deformation}

\def\mysubtitle{for Interactive Environments}
\def\myauthor  {Ben Kenwright} 
\def\myemail   {bkenwright@ieee.org (First Draft Feb 2015)} 
\def\mykeywords{vehicle, deformation, convex hulls, 3d, convex, video games, real-time computer generated, interactive}

\documentclass[conference,backref=page]{acmsiggraph}

\TOGonlineid{45678}
\TOGvolume{0}
\TOGnumber{0}
\TOGarticleDOI{1111111.2222222}
\TOGprojectURL{}
\TOGvideoURL{}
\TOGdataURL{}
\TOGcodeURL{}

\title{\mytitle \\ \fontsize{13}{16}\selectfont \mysubtitle}
\author{\myauthor\thanks{e-mail:\myemail} }
\pdfauthor{\myauthor}

\usepackage[compact]{titlesec}
\titlespacing{\section}{0pt}{0ex}{0ex}
\titlespacing{\subsection}{0pt}{0ex}{0ex}
\titlespacing{\subsubsection}{0pt}{0.0ex}{0ex}

\hypersetup{
    colorlinks  = true, 
    linkcolor   = blue,
    anchorcolor = red,
    citecolor   = blue, 
    filecolor   = red, 
    pdftex,
    pdfauthor={\myauthor},
    pdftitle={\mytitle},
    pdfsubject={\mytitle},
    pdfkeywords={\mykeywords},
    pdfcreator={\myauthor},
    pdfproducer={\myauthor},
}

\usepackage{enumitem}
\setitemize{leftmargin=13.0pt,noitemsep,topsep=0pt,parsep=0pt,partopsep=0pt}
\setenumerate{noitemsep,topsep=0pt,parsep=0pt,partopsep=0pt}

\usepackage{graphicx}

\graphicspath{{./images/}}

\newcommand{\figuremacroW}[4]{
	\begin{figure}[!htbp] 
		\centering
		\includegraphics[width=#4\columnwidth]{#1}
		\caption[#2]{\textbf{#2} - #3}
		\label{fig:#1}
	\end{figure}
}

\newcommand{\figuremacroF}[4]{
	\begin{figure*}[!htbp] 
		\centering
		\includegraphics[width=#4\textwidth]{#1}
		\caption[#2]{\textbf{#2} - #3}
		\label{fig:#1}
	\end{figure*}
}

\usepackage{amssymb}

\usepackage[ruled,linesnumbered]{algorithm2e}

\usepackage{setspace}

\begin{document}





\hypersetup{pdfinfo={
   Author		= {\myauthor},
   Title		= {\mytitle \mysubtitle},
   Subject 		= {\mytitle \mysubtitle},
   CreationDate = {D:20130530195600},
   Keywords 	= {\mykeywords},
}}

\pdfinfo{
   /Author (\myauthor)
   /Title  (\mytitle \mysubtitle)
   /Subject (\mytitle \mysubtitle)
   /CreationDate (D:20130530195600)
   /Keywords (\mykeywords)
}

\maketitle

\begin{abstract}
This paper proposes a real-time physically-based method for simulating vehicle deformation.  
Our system synthesizes vehicle deformation characteristics by considering a low-dimensional coupled vehicle body technique.  
We simulate the motion and crumbling behavior of vehicles smashing into rigid objects. 
We explain and demonstrate the combination of a reduced complexity non-linear finite element system that is scalable and computationally efficient.  We use an explicit position-based integration scheme to improve simulation speeds, while remaining stable and preserving modeling accuracy.  
We show our approach using a variety of vehicle deformation test cases which were simulated in real-time.

\end{abstract}

\section{Introduction}

\paragraph{Vehicle Deformation \& Damage}
Vehicle deformation is concerned with the object shape changing temporarily (elastic deformation) or permanently (plastic deformation or fracture) due to external forces, such as, impacts with the environment.
The simulation of realistic vehicle deformation in real-time is challenging and important owing to the complexity of the problem and the realistic engagement it provides.
When a vehicle impacts with an object, the vehicle does not bounce away like a rubber ball.  
While cars are design to be rigid and rugged, they are ultimately deformable for safety reasons.
The vehicle deforms and absorbs the impact energy during a crash so that most of the energy is dissipated across the body.
Hence, an effective and scalable technique for creating aesthetically pleasing vehicle deformation in real-time would be significant.
The area of damageable and destructible vehicles is a complex multi-discipline problem.  
Typically, a vehicle is decomposed of multiple materials (e.g., glass, rubber, plastic, and steel) which all deform or break in different manners.
For example, due to the large number of materials a vehicle is made from, this would cause a range of effects during an impact, such as, bending, tearing, splintering, and smashing.    However, for this paper, we focus on an uncomplicated deformation technique for real-time environments, such as, video games (e.g., cars bumping into objects, sliding along walls, and rolling, to produce emphasized dents and bends).

\paragraph{Aesthetically Pleasing}
The deformation focuses on \emph{entertainment appearances}, and does not mechanically affect the vehicle dynamics.
Adding dynamic interactive content to a simulation serves to enhance the enjoyability.
This paper does not strive to create an `ultra accurate simulation'.  Car damage in vehicle simulations (e.g., driving games) can add additional novelty and engagement without spoiling the focus (i.e., playing).  Vehicle deformation offers a dynamic challenge.
While it is common for commercial games to have soft-body physics, the question of doing it effectively and efficiently without sacrificing other features within the simulation.  Ideally, we want everything in the scene to react to external forces in a believable manner.  
For instance, in a racing simulation, we want to avoid a car going off track and rolling over, to recover without a scratch to continue on its journey.
Visually feeding back to the player through vehicle deformation collisions and damage aesthetically is valuable. 
Video games have issues putting car damage into their game because the modelling is difficult or the manufacturers of those cars will not let them depict what will happen if the player crashes their car into a wall, but can be a real disappointment when the player can  crash a vehicle into a barrier at 100+ mph and has nothing happens to it.
This paper presents an appropriately realistic damage/deformation model without too much work and without potentially causing detriment to the rest of the simulation (e.g., computational time and taking development resources away from other areas). 

\paragraph{Elastic and Plastic Deformation} 
A lot of real-time research has been done into creating \emph{elastic} deformable bodies \cite{capell2005physically,nesme2009preserving,kim2011fast} (e.g., cloth and soft body meshes).  This type of deformation is reversible. Once the forces are no longer applied, the object returns to its original shape.  Less work has addressed the \emph{plastic} deformation.
This type of deformation is irreversible. However, an object in the plastic deformation range will first have undergone elastic deformation, which is reversible, so the object will return part way to its original shape (see Figure \ref{fig:elasticplastic}).
The fracturing region is the final point after plastic deformation.  All materials will eventually fracture and break with sufficient force.  This type of deformation is also irreversible but we do not consider it in our approach. A break occurs after the material has reached the end of the elastic and plastic deformation ranges and would result in the mesh splitting and tearing.
Typically, a car body is majority made from steel.  As we would expect, steel has a small elastic deformation range and a rather large plastic deformation range, while materials, such as, plastics and rubber, have a minimal plastic deformation range and a larger elastic range. 
Research that has investigated elastic and plastic structural analysis of vehicles has primarily been off-line. These off-line models are able to synthesize accurate deformations \cite{zhang2008vehicle,moradi2013use} (e.g., elastic-plastic deformation) but are computationally expensive and not easily scalable or applicable to real-time environments.

\figuremacroW
{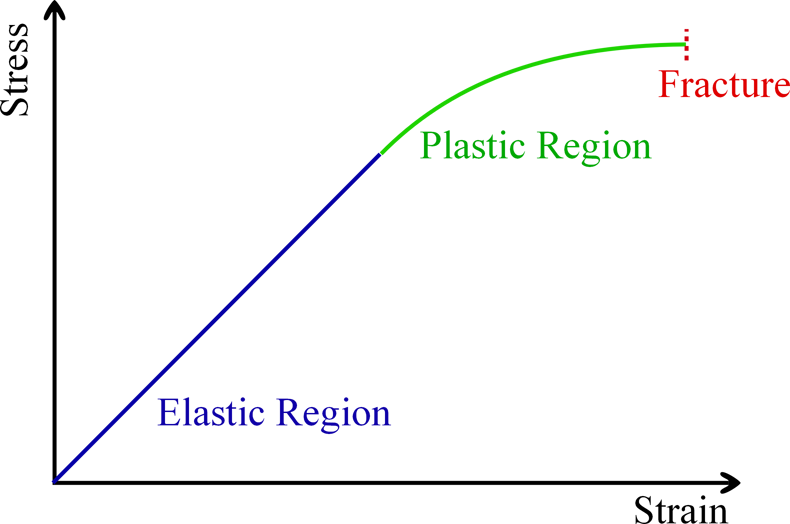}
{Elastic-Plastic Deformation}
{Illustrate the stress-strain curve and the relationship between stress (force applied) and strain (deformation).}
{0.6}
 
\paragraph{Contribution}
We paper presents a lightweight vehicle deformation system for real-time environments.
The simulation is fast enough for interactive systems, such as, vehicle-themed video games and driving simulators.  
The deformations generate in our system have the following features, and to our knowledge, no previous work has demonstrated them: 
(1) The simulation speed is fast enough for resource limited environments while maintaining an interactive frame-rate and dynamic model without sacrificing accuracy (e.g., the solution can be saved and shared over a network in real-time).
(2) Our system presents a unified solution for modeling vehicle damage characteristics (e.g., body-mass distribution) to capture life-like damage properties.
(3) All the computational and memory run-time costs for applying and solving the deformation can be done in a single time-step (i.e., a linearly scalable model).

 \section{Related Work}

The creation and control of deformable vehicle meshes is an important component in multiple research areas, such as, engineering material analysis and safety testing.  However, we focus on an interactive solution, such as, video games, since it allows the player to visually experience the effect of real-life car accidents.
We review a number of important papers that have presented solutions for deformation in different contexts (e.g., character animation and material analysis) that relate to our work.

\figuremacroW
{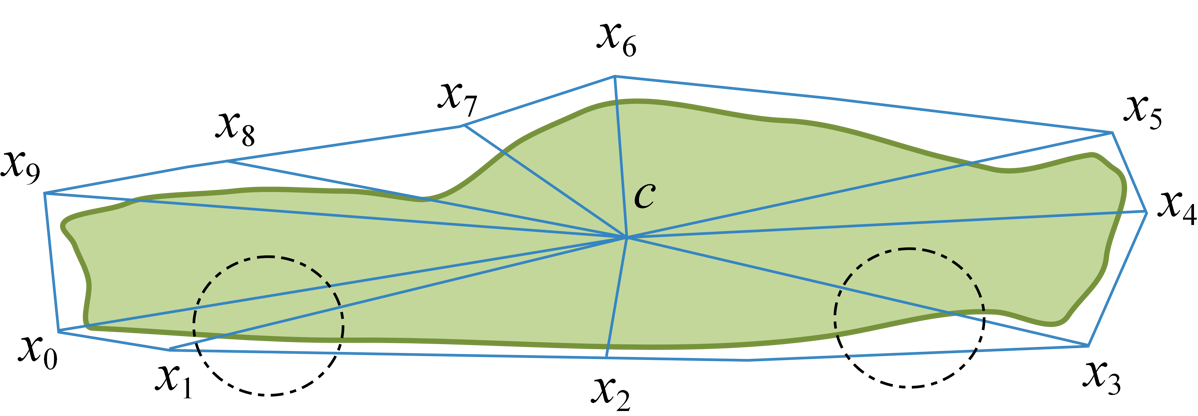}
{Concept}
{A 2D illustration of the low-dimensional deformable body driven system.
The course volumetric mesh encloses the detailed graphical surface.  
The deformable body is attached to the graphical mesh by nodes to produce the overall effect. 
In this illustration, $c$ is the centroid and $x_0, .. ,x_n$ represents the coordinates for the coarse deformable mesh.
}
{0.8}

While accurate simulations of vehicle deformation have been published \cite{moradi2013use}, a real-time solution that is scalable and flexible has not yet been presented.
Similarly, in commercial circles (e.g., video games), we see vehicle deformation and damage but the method and techniques are propriety owned and not shared publically.
One open-source vehicle damage system is available, known as `Rigs of Rods (RoR)' \cite{rigsofrods}, which uses full soft-body physics to emulate the structural breakdown of vehicles but can be computationally intensive.  The framework models the vehicle characteristics by accurately decomposing the model chassis into the correct components with stress characteristics to produce a highly realistic solution with one goal (vehicle destruction).  
Our approach sees deformation and damage as a minor part of the system, since a real-world solution only provides a minute amount of the overall computational resources to the physics - needing to share with graphics, artificial intelligence, networking, and game-play features.

A method that does not encapsulate the physical properties of the model, but is able to produce uncomplicated vehicle deformation feedback uses texture-vertex mapping.  
The approach writes collisions to a texture that is used to deform the graphical mesh vertices (e.g., analogous to physical bump-mapping).
Information written to the texture are applied to the vehicle mesh using the graphical processing unit (GPU) to achieve a real-time frame-rates.  
This approach can also be used to write non-deformable feedback (e.g., scratches and marks to the graphical textures).

\section{Deformation Model}
This section discusses the representation of the deformation model (i.e., finite element decomposition approximation).
The model must be able to handle large deformations and be stable under large time-steps while not hindering the systems performance.
Deformation is a geometric measure of strain (e.g., stretching and shearing).
Typically, strain models use accurate finite element methods that assume small deformations.  These simple linear strain models may cause inflation/expansion issues when the strain cannot separate rotation information \cite{muller2004interactive}.  Of course, methods have been developed to attempt to remove the rigid rotation \cite{capell2002multiresolution,terzopoulos1987elastically}.
This solution is popular in interactive applications but the deformation needs to be very small.
We use a coarser non-linear finite-element model that is coupled to the high-detailed geometry mesh.  This allows us to synthesize physically plausible deformation in real-time, while maintaining a reasonably correct physical model. Our coarser model extracts the key details from the high-detailed vehicle mesh, such as, structural inter-connectivity and mass distribution.

Vehicle collision deformations are likely to be large, especially for high speed impacts (e.g., car-car collisions).
We must take care with these large deformations, as large aesthetically pleasing deformation are difficult to create in some respects (i.e., materials should not look rubbery or jelly-like).
In order to handle large deformations, we use a low-dimensional approximation.  We express large deformations regardless of vehicle rigid body centre (i.e., the vehicle dynamics).  As the vehicle dynamics are not by default hindered by the deformation.
From the viewpoint of computational speed, using a lightweight strain approximation with a reduced number of mesh elements (i.e., a coarser mesh based on the vehicle's convex hull, Section \ref{sec:lowmesh}) that is parallelizable, makes it possible to produce acceptable large deformations at real-time simulation speeds.

The mesh node positions formulate a finite element decomposition.  The strain for each element is computed from the nodal positions in the currently deformed state and the initially non-deformed state. 
For every element, we determine the influence of each neighbour.  As nodes are disturbed from their rest location by external perturbations they influence their neighbouring elements in accordance with their spatial proximity and connectivity strength.

\figuremacroW
{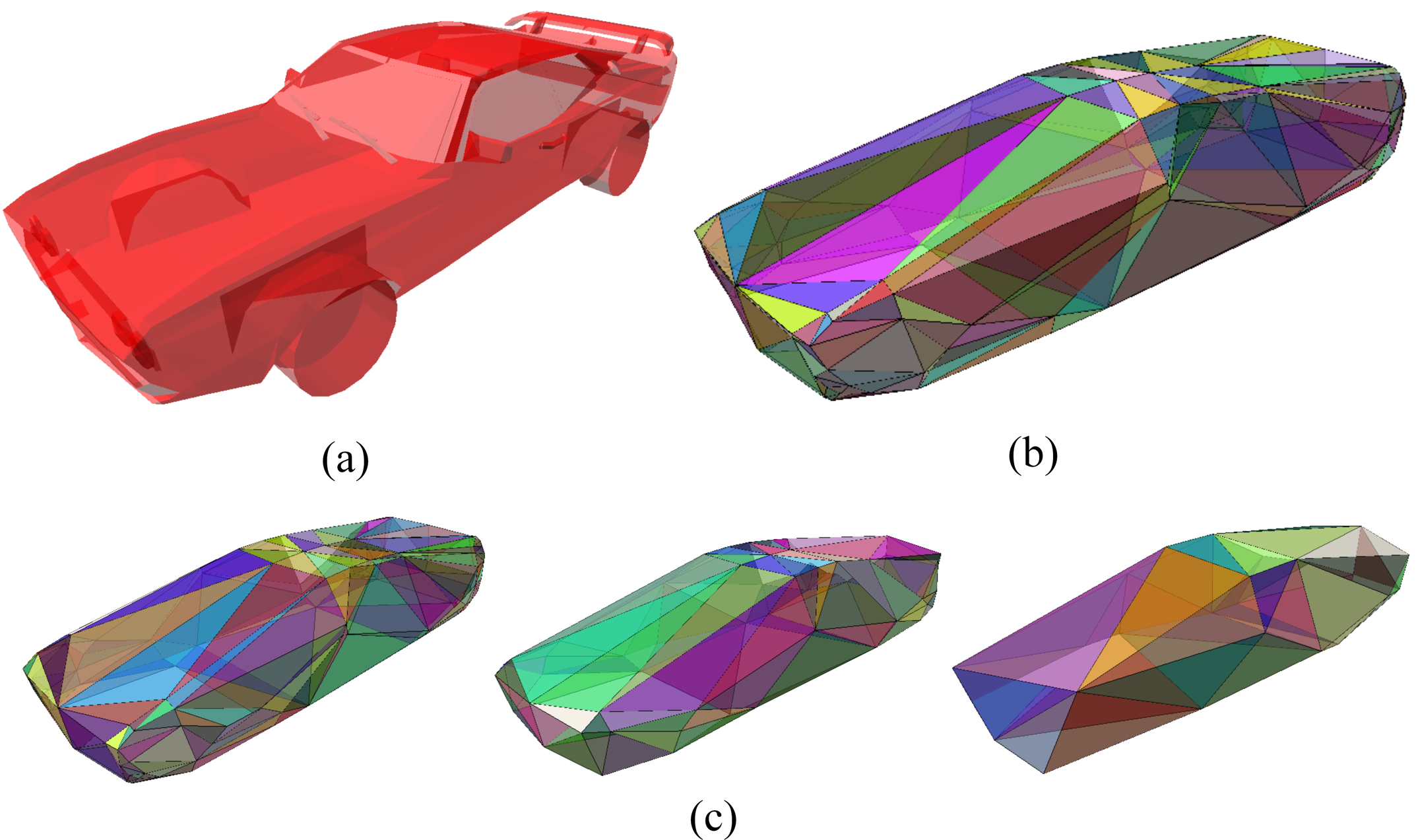}
{Vehicle Convex Hull}
{Generating a convex hull mesh for the vehicle model. (a) original mesh, (b) convex hull for collisions and control points, and (c) different levels of detail for the deformation control model by reducing the convex hull triangles and hence number of control points.}
{1.0}

\figuremacroW
{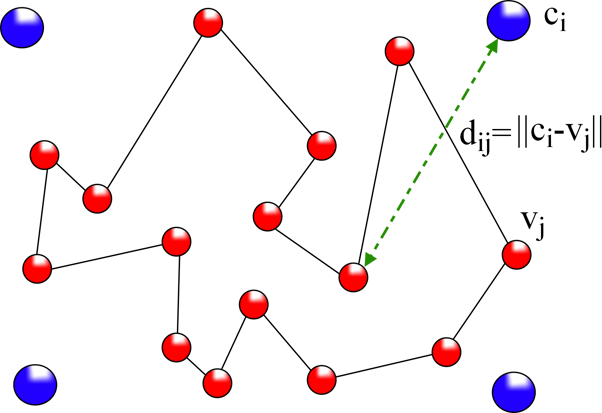}
{Control Points and Vertices}
{The control points coordinate the deformation of the high-detail graphical mesh.  So in Equation \ref{eq:meshcontrol}, we take the inverse distance, so that nearer the point the greater the influence.  Taking the inverse power enables us to regionalize the influence, due to the exponential fall off and provide a stronger coupling, between the graphical mesh and the control points (see Figure \ref{fig:controlpoints}.}
{0.6}

\section{Low-Resolution Control Mesh} \label{sec:lowmesh}
We explain our low-dimensional control model for vehicle deformation.
A control mesh technique is used reduce the computational overhead and the mathematical complexity of the model so we can achieve real-time frame-rates. 
We explain the high-resolution vehicle surface interactions to handle detailed contacts between the
vehicle and the virtual environment in an endeavour to realistically mimic the mechanical deformation properties.

\figuremacroF
{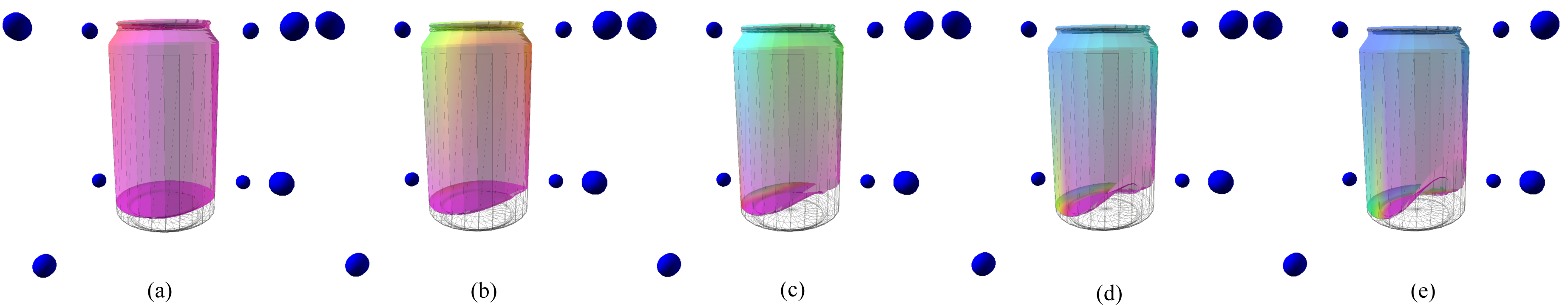}
{Control Point Weighting}
{Example of the biasing influences of control points to different powers (a) $\alpha=1$, (b) $\alpha=2$, (c) $\alpha=3$, (d) $\alpha=4$, and (e) $\alpha=5$.  Increasing the inverse distance to the power means the control point only influences points within its region.  For example, in (a) the whole mesh is influenced by moving the corner control point upwards, while in (e) only the corner points are influence.  Additionally, the control points have a stronger connection.}
{1.0}

\paragraph{Model Reduction \& Mesh Embedding}

The two main techniques for reducing the complexity of a finite element system can be classified into two main types: 
\emph{modal reduction} and \emph{mesh embedding}. Modal reduction is a popular method for reducing the complexity of a finite element system by using a linear subspace to span a small number of displacement basis vectors to represent the deformation in the body. 
The eigenmodes obtained from linear modal analysis would be the best basis vectors for small deformation. 
For large deformation, however, they are not sufficient to capture the non-linear deformation characteristics, so multiple techniques have been suggested to choose a good deformation basis set \cite{barbivc2005real}.
Model techniques have successfully been used for
real-time solutions, such as, surgery
simulators and hand-soft body interaction.

Mesh embedding, which is also called \emph{free-form} deformation \cite{kim2011fast}, uses a low-dimensional coarse volumetric mesh to enclose the entire deformable body in order to represent the behavior of the body. The location of every material point inside the deformable body is determined by interpolating the positions of the neighboring nodes in the mesh. 
Since the work by Faloutsos et al. \cite{faloutsos1997dynamic}, mesh embedding techniques have been widely used to simulate soft bodies in the graphics literature \cite{nesme2009preserving,kharevych2009numerical,capell2005physically}
We chose mesh embedding to reduce complexity of the deformable body in our simulation system not only because the technique can reduce the model complexity without losing the fine geometry of the vehicle but also because the frame can be manipulated more easily and efficiently using the embedding mesh system compared to modal reduction. In our formulation, the control body elements are considered as an interconnected set of rigid elements that can be solved using iterative penalty-based constraints. 
The complete system consists of a set of deformable body elements and a rigid body core (i.e., vehicle centre-of-mass).

The position of a material point in the deformable body is determined from the nodal positions of the coarse mesh through interpolation.  The relationship between the vehicle mesh and the nodes is defined in Equation \ref{eq:meshcontrol}.

\begin{equation} \label{eq:meshcontrol}
\begin{alignedat}{3}
\phi_{ij} &= \frac{1}{|| c_i - v_j ||^\alpha} \\
v'_j      &= v_j - \left[ \sum_i(c_{i0} \; \phi_{ij}) - \sum_i(c_{i} \phi_{ij}) \right] \\
\end{alignedat}
\end{equation}

\noindent  where 
		   $v_j$ is the $j$th vertex , 
		   $c_i$ is the $i$th control point,
		   $v'_j$ is the $j$th transformed output vertex,
		   $\phi_{ij}$ is the initial inverse distance between vertex $j$ and control point $i$ (is constant and calculated once at the beginning), and 
		   $\sum_i \phi_i = 1$ (typically, $\alpha \approx 3-4$).

\paragraph{Scalability}
We endeavour to automate the damage process rather than depending on artist intervention for modelling the underlying low-poly control mesh.
This enables us to adapt the detail of the deformation model to scale to different platforms (i.e., reduce the model complexity to more coarser representations for environments with limited resources, such as, memory and processing power).

\figuremacroW
{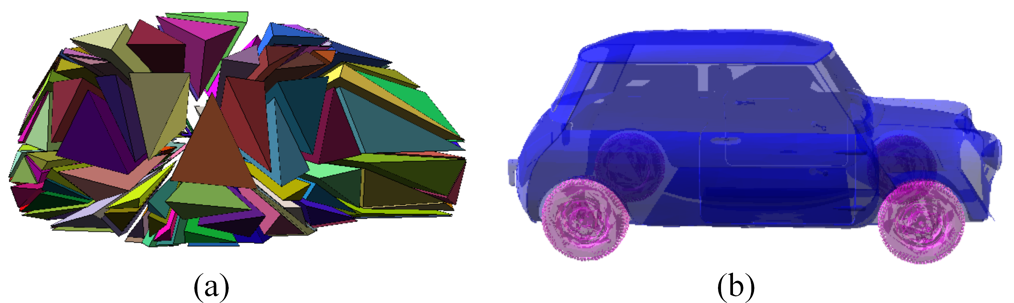}
{Non-Linear Control Element Coupling}
{Decompose the coarse mesh into tetrahedrons to calculate the mass distribution for each element (a) exploded view of the convex hull, and (b) original high resolution mesh.}
{1.0}

\section{Dynamics}
In order to capture the structural failing of a vehicle body during impacts, the constraints are assumed to break when their deformation exceeds a certain threshold value.  Under loading, the vehicle body will suffer from two forms of deformation, i.e., elastic and plastic.  The body shell is characterized to bend and return to its original shape for small disturbances.  When the elastic threshold is exceeded the body will deform and take on a new shape (i.e., a new rest shape).

Linear elastic deformation is governed by Hooke`s law, which states, $\sigma = E \varepsilon$, where $\varepsilon$ is the strain, $\sigma$ is the applied stress, and $E$ is a material constant called Young`s modulus. 
Importantly, this relationship \emph{only applies in the elastic range} and indicates that the slope of the stress vs. strain curve can be used to find Young`s modulus. 
In engineering, this calculation is used to determine the materials tensile strength \cite{callister2001fundamentals}. 
The elastic range ends when the material reaches its yield strength. At this point plastic deformation begins.

The deformable shell constraints are enforced using position-based dynamics.  This provides control over the explicit integration and remove instability issues.
We were able to manipulate the positions of the vertices and parts of deformable mesh directly during the simulation without complications.
Our approach is formulated to handle general position-based constraints effortlessly and efficiently.  Additionally, an explicit position-based constraint solver is easy to understand and implement.

We define general constraints via a constraint function (\cite{teschner2004versatile,muller2007position,baraff1998large}). 
Instead of computing forces as the derivative of a constraint function energy, we directly solve for the equilibrium configuration and project positions.
With our method we derive a bending term for the material which uses a point based approach  similar to the one proposed by Grinspun et al. \cite{grinspun2003discrete} and Bridson et al. \cite{bridson2003simulation}.

Position-based dynamics have been used for a variety of systems.  For example, Jakobsen \cite{jakobsen2001advanced} built his physics engine (called Fysix) on a position-based approach. With the central idea of using a Verlet integrator to manipulate positions directly.  The velocities are implicitly stored by the current and the previous positions of the point-masses.  The constraints are enforced by direct manipulation of the positions. Jakobsen demonstrated the efficient representation fo distance constraints that could be used to form the underpinnings of a stable and iterative control mesh. 
In this paper, we use position-based constraints with a breaking threshold, which is used to create a semi-rigid shell for the vehicle.  
After breaking the constraint rest conditions are recalculated to represent the new positions.
An important note is our model is decomposed of point masses and does not need to account for any angular calculations.
Position-based methods have proven themselves an efficient and robust method in variety of soft body systems, such as, cloth \cite{muller2007position}, character animation \cite{jakobsen2001advanced}, and fluid dynamics \cite{macklin2013position}.  For a detailed introduction to position-based dynamics, we refer the reader to the interesting work of M{\"u}ller \cite{muller2007position} and Jakobsen \cite{jakobsen2001advanced}.

\paragraph{Visual Constraints}
We set limits on the amount of deformation (i.e., deviation between the starting and current control point locations).  The control points were only allowed to deviate by a specified amount from their starting locations.  This was set as a global constant for the vehicle, however, it could be customized for different control points to produce a more aesthetically correct result if necessary.

\section{Experimental Results}
We focused on three-dimensional vehicle deformation, however, we also applied the concept to test models to emphasis particular characteristics (e.g., a drinks can shown in Figure \ref{fig:controldetails}).
Various kinds of simulations were tested in our system with the three-dimensional deformable vehicle bodies shown in Figure \ref{fig:crashintowall}.
The simulations with all the test models were implemented on a desktop machine with 3.2 GHz
Intel i7 CPU and NVIDIA GeForce GTX 480 GPU. 

In order to create physically plausible vehicle motions (i.e., a car driven by its wheels), we used only a basic rigid body simulator, applying external forces at the locations of the wheels to  acceleration the body, and drive the overall vehicle motion.
The secondary motion of the deformable mesh were obtained by coupling the control points to the rigid body chassis.  The deformations cause cause the rigid body collision mesh to be updated each time the deformation mesh is modified, for instance, by external forces, such as, collisions and contacts, to produce the desired visual damage.

\paragraph{Bottlenecks}
Typically, a vehicle deformation model can be difficult to simulate due to a number of challenges:
\begin{itemize}
\item model complexity (e.g., number of triangles and sub-mesh objects, such as, doors, chairs, engine, and windows)
\item inter body collisions (e.g., mesh-mesh interaction) - typically for interactive environments, we are not interested in the internal modelling of the vehicle (e.g., engine, suspension, AND impact bars).  Hence, during collisions and deformations, we avoid showing the internal body since it this information is not provided by the artist (i.e., aesthetic outer model of the car) 
\item interconnected physical constraints (e.g., formulating complex matrix model to realistically connect constraints)
\item crumpling and compression factors (i.e., certain materials deform in specialist ways, such as, folding and crumpling)
\end{itemize}

\section{Discussion}
This paper focuses on an aesthetically pleasing vehicle deformation model for real-time interactive environments.  Realistic vehicles are designed with specialist deformation zones to help protect the passengers in the event collisions. 
Typically, simulation developers are limited by numerous issues, both from resources and from legally.  For example, video game publishers/developers who use real-world vehicle models need permission from car manufacturers to what damage they show within their virtual environment.
Of course, deformation simulates interactive effects and provides a more engaging virtual world for the user.
We presented vehicle deformation framework for the vehicle and not the environments, for instance, collisions with barriers, which would also need to be deformed the barrier for additional realism. 
We created a scalable and effective method for creating deformation in real-time, which can be expanded upon to incorporate additional damage features, such as, fracturing and tearing of the vehicle body.
The model is flexible enough to be expand to include greater complexity, such as, different material plasticity and elasticity constants (e.g., bumper and bonnet).  While we have used an uncomplicated point-mass decomposition with linear coupling between the low and high resolution mesh, additional methods from skinning \cite{kavan2007skinning,kenwright2012beginners}, such as, dual-quaternion interpolation, may offer smoother surfaces with bulging characteristics.

\section{Conclusion}

A robust and efficient method for adding deformations vehicle simulation environments produces a more engaging and entertaining solution.  The method we have presented allows the solution to be customized to the systems needs (e.g., scalable, efficient, and autonomous or through artistic customisation).
The flexibility of the approach within this paper allows developers and artists to design more attractive vehicle simulations that are more active and engaging without sacrificing resources.
Overall, we focused on a low-dimensional solution which moves away from pre-canned solutions (i.e., stored animation files) and supports a diverse set of characteristics to create an effective real-time effect.

\section*{Acknowledgements}
A special thanks to reviewers for taking time to review this article and provide insightful comments and suggestions to help to improve the quality of this article.

\figuremacroW
{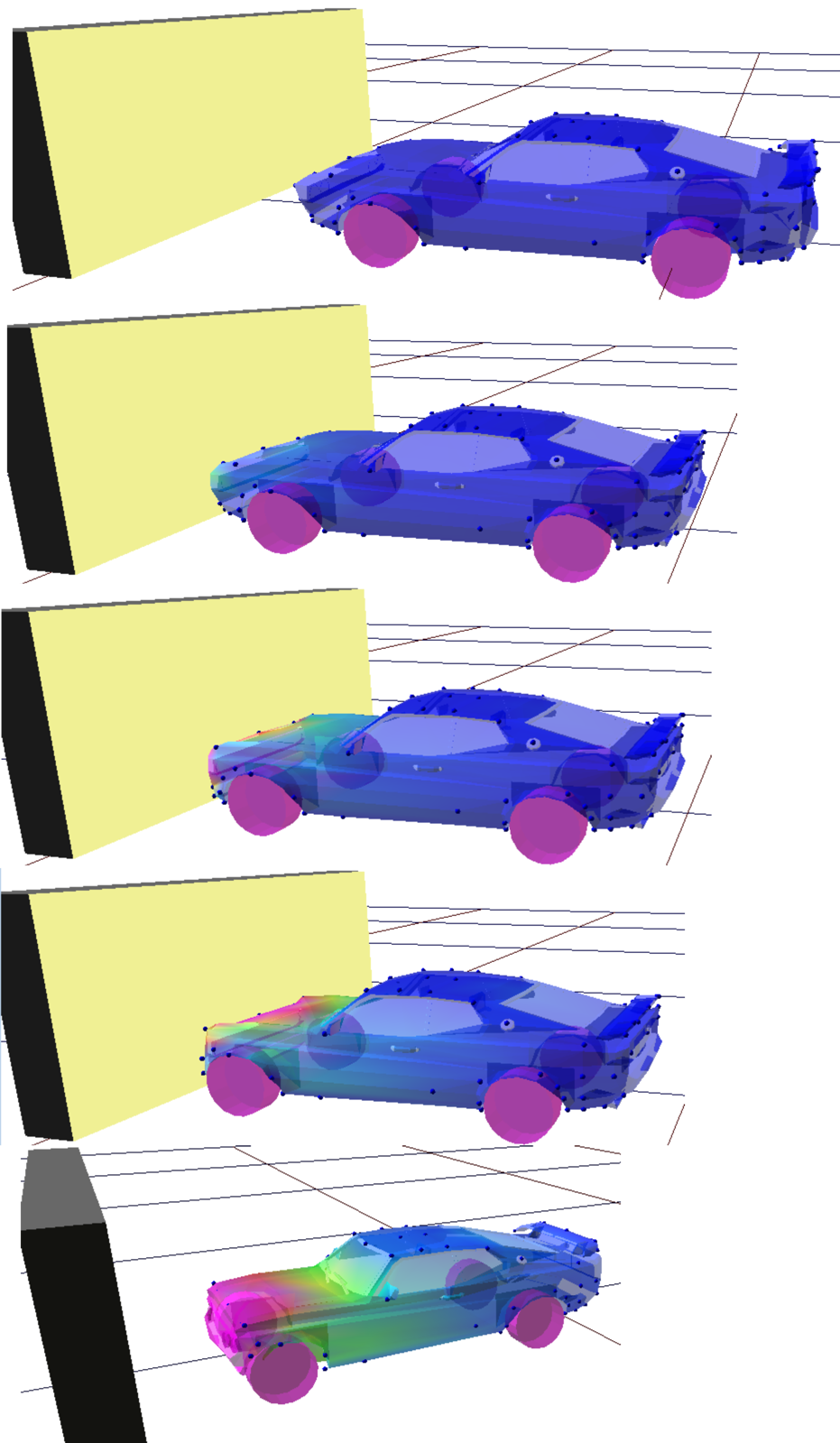}
{Direct Wall Collision}
{An example of a head on crash into a flat wall. We can visually see the spatial influence of the front deformation across the car's body.
Emphasising that a vehicle distributes a collision like a wave across the vehicle. 
The deformation travels across the car like ripples on water. 
We can visualize the coupled influence of the deformation across the car using colors (i.e., red largest influence and blue no influence).
}
{1.0}

\figuremacroW
{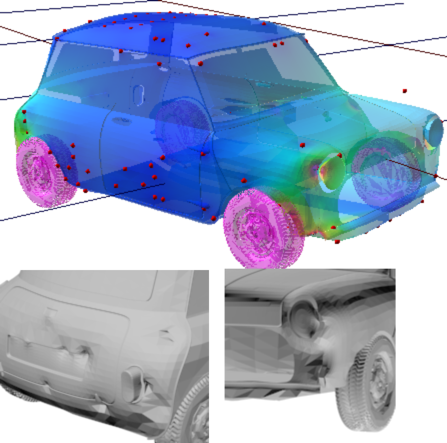}
{Mesh Deformation}
{The deformation of the vehicle mesh creates a more original experience to individualize each player's car.  A vehicle with 900,000 vertices and 60 control points that has bumped into minor objects (e.g., posts and vegetation).
}
{1.0}


\bibliographystyle{acmsiggraph}



\let\oldthebibliography=\thebibliography
\let\endoldthebibliography=\endthebibliography

\renewenvironment{thebibliography}[1]{%
    \begin{oldthebibliography}{#1}%
      \setlength{\parskip}{0ex}%
      \setlength{\itemsep}{0.5ex}%
  }%
  {%
    \end{oldthebibliography}%
  }


\bibliography{paper}

\def\url#1{}
\begin{thebibliography}{\protect\citename{Terzopoulos et~al\mbox{.} }1987}

\bibitem[\protect\citename{Baraff and Witkin }1998]{baraff1998large}
{\sc Baraff, D., and Witkin, A.}
\newblock 1998.
\newblock Large steps in cloth simulation.
\newblock In {\em Proceedings of the 25th annual conference on Computer
  graphics and interactive techniques}, ACM, 43--54.

\bibitem[\protect\citename{Barbi{\v{c}} and James }2005]{barbivc2005real}
{\sc Barbi{\v{c}}, J., and James, D.~L.}
\newblock 2005.
\newblock Real-time subspace integration for st. venant-kirchhoff deformable
  models.
\newblock In {\em ACM Transactions on Graphics (TOG)}, vol.~24, ACM, 982--990.

\bibitem[\protect\citename{Bridson et~al\mbox{.} }2003]{bridson2003simulation}
{\sc Bridson, R., Marino, S., and Fedkiw, R.}
\newblock 2003.
\newblock Simulation of clothing with folds and wrinkles.
\newblock In {\em Proceedings of the 2003 ACM SIGGRAPH/Eurographics symposium
  on Computer animation}, Eurographics Association, 28--36.

\bibitem[\protect\citename{Callister }2001]{callister2001fundamentals}
{\sc Callister, W.~D.}
\newblock 2001.
\newblock Fundamentals of materials science and engineering.

\bibitem[\protect\citename{Capell et~al\mbox{.}
  }2002]{capell2002multiresolution}
{\sc Capell, S., Green, S., Curless, B., Duchamp, T., and Popovi{\'c}, Z.}
\newblock 2002.
\newblock A multiresolution framework for dynamic deformations.
\newblock In {\em Proceedings of the 2002 ACM SIGGRAPH/Eurographics symposium
  on Computer animation}, ACM, 41--47.

\bibitem[\protect\citename{Capell et~al\mbox{.} }2005]{capell2005physically}
{\sc Capell, S., Burkhart, M., Curless, B., Duchamp, T., and Popovi{\'c}, Z.}
\newblock 2005.
\newblock Physically based rigging for deformable characters.
\newblock In {\em Proceedings of the 2005 ACM SIGGRAPH/Eurographics symposium
  on Computer animation}, ACM, 301--310.

\bibitem[\protect\citename{Faloutsos et~al\mbox{.} }1997]{faloutsos1997dynamic}
{\sc Faloutsos, P., Van De~Panne, M., and Terzopoulos, D.}
\newblock 1997.
\newblock Dynamic free-form deformations for animation synthesis.
\newblock {\em Visualization and Computer Graphics, IEEE Transactions on 3}, 3,
  201--214.

\bibitem[\protect\citename{Grinspun et~al\mbox{.} }2003]{grinspun2003discrete}
{\sc Grinspun, E., Hirani, A.~N., Desbrun, M., and Schr{\"o}der, P.}
\newblock 2003.
\newblock Discrete shells.
\newblock In {\em Proceedings of the 2003 ACM SIGGRAPH/Eurographics symposium
  on Computer animation}, Eurographics Association, 62--67.

\bibitem[\protect\citename{Jakobsen }2001]{jakobsen2001advanced}
{\sc Jakobsen, T.}
\newblock 2001.
\newblock Advanced character physics.
\newblock In {\em Game Developers Conference}, 383--401.

\bibitem[\protect\citename{Kavan et~al\mbox{.} }2007]{kavan2007skinning}
{\sc Kavan, L., Collins, S., {\v{Z}}{\'a}ra, J., and O'Sullivan, C.}
\newblock 2007.
\newblock Skinning with dual quaternions.
\newblock In {\em Proceedings of the 2007 symposium on Interactive 3D graphics
  and games}, ACM, 39--46.

\bibitem[\protect\citename{Kenwright }2012]{kenwright2012beginners}
{\sc Kenwright, B.}
\newblock 2012.
\newblock A beginners guide to dual-quaternions: what they are, how they work,
  and how to use them for 3d character hierarchies.

\bibitem[\protect\citename{Kharevych et~al\mbox{.}
  }2009]{kharevych2009numerical}
{\sc Kharevych, L., Mullen, P., Owhadi, H., and Desbrun, M.}
\newblock 2009.
\newblock Numerical coarsening of inhomogeneous elastic materials.
\newblock In {\em ACM Transactions on Graphics (TOG)}, vol.~28, ACM, 51.

\bibitem[\protect\citename{Kim and Pollard }2011]{kim2011fast}
{\sc Kim, J., and Pollard, N.~S.}
\newblock 2011.
\newblock Fast simulation of skeleton-driven deformable body characters.
\newblock {\em ACM Transactions on Graphics (TOG) 30}, 5, 121.

\bibitem[\protect\citename{Macklin and M{\"u}ller }2013]{macklin2013position}
{\sc Macklin, M., and M{\"u}ller, M.}
\newblock 2013.
\newblock Position based fluids.
\newblock {\em ACM Transactions on Graphics (TOG) 32}, 4, 104.

\bibitem[\protect\citename{Moradi et~al\mbox{.} }2013]{moradi2013use}
{\sc Moradi, R., Setpally, R., and Lankarani, H.~M.}
\newblock 2013.
\newblock Use of finite element analysis for the prediction of driver fatality
  ratio based on vehicle intrusion ratio in head-on collisions.
\newblock {\em Applied Mathematics 4\/}, 56.

\bibitem[\protect\citename{M{\"u}ller and Gross }2004]{muller2004interactive}
{\sc M{\"u}ller, M., and Gross, M.}
\newblock 2004.
\newblock Interactive virtual materials.
\newblock In {\em Proceedings of Graphics Interface 2004}, Canadian
  Human-Computer Communications Society, 239--246.

\bibitem[\protect\citename{M{\"u}ller et~al\mbox{.} }2007]{muller2007position}
{\sc M{\"u}ller, M., Heidelberger, B., Hennix, M., and Ratcliff, J.}
\newblock 2007.
\newblock Position based dynamics.
\newblock {\em Journal of Visual Communication and Image Representation 18}, 2,
  109--118.

\bibitem[\protect\citename{Nesme et~al\mbox{.} }2009]{nesme2009preserving}
{\sc Nesme, M., Kry, P.~G., Je{\v{r}}{\'a}bkov{\'a}, L., and Faure, F.}
\newblock 2009.
\newblock Preserving topology and elasticity for embedded deformable models.
\newblock In {\em ACM Transactions on Graphics (TOG)}, vol.~28, ACM, 52.

\bibitem[\protect\citename{ROR }2012]{rigsofrods}
{\sc ROR}, 2012.
\newblock Rigs of rods (ror) - {(http://www.rigsofrods.com) Accessed:
  02/01/2015}.

\bibitem[\protect\citename{Terzopoulos et~al\mbox{.}
  }1987]{terzopoulos1987elastically}
{\sc Terzopoulos, D., Platt, J., Barr, A., and Fleischer, K.}
\newblock 1987.
\newblock Elastically deformable models.
\newblock In {\em ACM Siggraph Computer Graphics}, vol.~21, ACM, 205--214.

\bibitem[\protect\citename{Teschner et~al\mbox{.} }2004]{teschner2004versatile}
{\sc Teschner, M., Heidelberger, B., Muller, M., and Gross, M.}
\newblock 2004.
\newblock A versatile and robust model for geometrically complex deformable
  solids.
\newblock In {\em Computer Graphics International, 2004. Proceedings}, IEEE,
  312--319.

\bibitem[\protect\citename{Zhang et~al\mbox{.} }2008]{zhang2008vehicle}
{\sc Zhang, X.-y., Jin, X.-l., Qi, W.-g., and Guo, Y.-z.}
\newblock 2008.
\newblock Vehicle crash accident reconstruction based on the analysis 3d
  deformation of the auto-body.
\newblock {\em Advances in Engineering Software 39}, 6, 459--465.

\end{thebibliography}

\end{document}